\documentclass{article}
\usepackage{cite}
\usepackage{amsmath,amssymb,amsfonts}
\usepackage{algorithmicx}
\usepackage[ruled]{algorithm2e}
\usepackage{graphicx}
\usepackage{subfigure}
\usepackage{epstopdf}
\usepackage{caption}
\usepackage{textcomp}
\usepackage{booktabs}
\usepackage{multirow}

\title{Tensor $p$-shrinkage nuclear norm for low-rank tensor completion}
\author{Chunsheng Liu, Hong Shan, Chunlei Chen}

\begin{document}
	\maketitle
\begin{abstract}
In this paper, a new definition of tensor $p$-shrinkage nuclear norm ($p$-TNN) is proposed based on tensor singular value decomposition (t-SVD). In particular, it can be proved that $p$-TNN is a better approximation of the tensor average rank than the tensor nuclear norm when $-\infty < p < 1$. Therefore, by employing the $p$-shrinkage nuclear norm, a novel low-rank tensor completion (LRTC) model is proposed to estimate a tensor from its partial observations. Statistically, the upper bound of recovery error is provided for the LRTC model. Furthermore, an efficient algorithm, accelerated by the adaptive momentum scheme, is developed to solve the resulting nonconvex optimization problem. It can be further guaranteed that the algorithm enjoys a global convergence rate under the smoothness assumption. Numerical experiments conducted on both synthetic and real-world data sets verify our results and demonstrate the superiority of our $p$-TNN in LRTC problems over several state-of-the-art methods.  
\end{abstract}

\section{Introduction}
\label{sec:introduction}
In the fields of computer vision and signal processing, there are massive multi-dimensional data that needs to be analyzed and processed. In particular, multi-dimensional arrays (i.e., tensors) provide a natural representation form for these data. Tensor, which is regarded as a multi-linear generalization of matrix/vector, can mathematically model the multi-dimensional data structures, making tensor learning so attractive that there are increasing applications in computer vision \cite{b1,b2,b3,b4}, machine learning \cite{b5,b6}, signal processing \cite{b7,b8}, and pattern recognition \cite{b9} in recent years. Unfortunately, the challenge of missing elements in the actually observed tensors limits its applications. Inspired by matrix completion \cite{b10}, tensor completion attempts to recover the underlying tensor, that on the low-rank assumption, from its incomplete observations. Mathematically, the low-rank tensor completion (LRTC) problem can be formulated as the following model:
\begin{equation}
\mathop {\min }\limits_{{\cal X}} rank\left( {{\cal X}} \right){\kern 1pt} {\kern 1pt} {\kern 1pt} {\kern 1pt} {\kern 1pt} {\kern 1pt} {\kern 1pt} {\kern 1pt} {\kern 1pt} {\kern 1pt} {\kern 1pt} s.t.{{{\cal P}}_\Omega }\left( {{\cal X}} \right) = {{{\cal P}}_\Omega }\left( {{\cal T}} \right).\label{eq}
\end{equation}
where ${{\cal X}},{{\cal T}} \in {^{{I_1} \times  \cdots  \times {I_n}}}$ denotes the underlying tensor and its observation tensor, respectively, $rank\left( {{\cal X}} \right)$ denotes the rank function of  ${\cal X}$, $\Omega$  denotes the index set of observed entries, and  ${{{\cal P}}_\Omega }\left( {{\cal X}} \right)$ denotes the projection operator that ${\left[ {{{{\cal P}}_\Omega }\left( {{\cal X}} \right)} \right]_{{i_1} \cdots {i_n}}} = {{{\cal X}}_{{i_1} \cdots {i_n}}}$ if $\left( {{i_1}, \cdots ,{i_n}} \right) \in \Omega $ and 0 otherwise. Different from matrix case, however, the definition of the tensor rank is not unique, and we can see that each definition corresponds to the particular tensor decomposition \cite{b11}. For example, tensor n-rank \cite{b12} is related to the Tucker decomposition \cite{b13}, CP-rank \cite{b14} is related to the CANDECOMP/PARAFAC (CP) decomposition \cite{b15}, tensor train (TT) rank is related to the TT decomposition \cite{b16}, and tubal rank \cite{b2} and tensor multi-rank are related to tensor singular value decomposition (t-SVD) \cite{b17}.

Due to the non-convex and non-smoothness of the rank function $rank\left(  \cdot  \right)$, the calculation of it is usually NP-hard problem that cannot be solved within polynomial time. Therefore, the function $rank\left( {{\cal X}} \right)$ is usually relaxed as its convex/non-convex surrogate, and \eqref{eq} can be rewritten as follows:
\begin{equation}
\mathop {\min }\limits_{{\cal X}} \ell \left( {{\cal X}} \right) + \lambda f\left( {{\cal X}} \right).\label{eq2}
\end{equation}
where $f$ denotes the surrogate function, $\lambda$ denotes the regularization parameter, and $\ell$ denotes the smooth loss function. Obviously, the difference among the existing LRTC models mainly lies in the choice of $f$. Although various tensor rank surrogates \cite{b3,b6,b18} are proposed to approximate the tensor rank, they all face some challenges in practical applications.

According to the CP decomposition \cite{b15}, Friedland et al. \cite{b18} claimed that the CP-rank could be relaxed it with CP tensor nuclear norm (CNN), which is a convex surrogate that can be defined as:
\begin{equation}
{\left\| {{\cal X}} \right\|_{CNN}} = \inf \left\{ {\sum\limits_{i = 1}^r {\left| {{\lambda _i}} \right|} \left| {{{\cal X}} = \sum\limits_{i = 1}^r {{\lambda _i}{\mathbf{u}_{1,i}} \circ {\mathbf{u}_{2,i}} \circ  \cdots  \circ {\mathbf{u}_{n,i}}} } \right.} \right\}.\label{eq3}
\end{equation}
where ${{\cal X}} \in {^{{I_1} \times  \cdots  \times {I_n}}}$, $\left\| {{\mathbf{u}_{j,i}}} \right\| = 1,j = 1, \cdots ,n$, and $\circ$ denotes the outer product. Moreover, some mathematical properties of CNN are also presented. Under the framework of Frank-Wolfe method, Yang et al. \cite{b19} directly relaxed the CP-rank with CNN and pointed out that the proposed model can be solved by converting it to calculate the corresponding spectral norm and rank-one tensors. In addition, Yuan et al. \cite{b20} developed the sub-differential of CNN and attempted to recover the underlying tensor with a dual certificate.

Although the CNN minimization defined above is convex, any efficient implementation has not been developed so far. Indeed, it is generally NP-complete to compute the CP-rank of tensor, as well as the tensor spectral norm \cite{b21}. Furthermore, the tightness of CNN relative to the CP-rank is hard to measure. Therefore, the application of CNN in LRTC problem is limited.

To avoid the NP-complete CP-rank calculations, tensor n-rank \cite{b12} is another commonly used tensor rank. For a given tensor ${{\cal X}} \in {^{{I_1} \times  \cdots  \times {I_n}}}$, the tensor n-rank of $\cal X$ is defined as ${{rank}_{tc}}\left( {{\cal X}} \right) = \left( {rank\left( {{\mathbf{X}_{\left( 1 \right)}}} \right), \cdots ,rank\left( {{\mathbf{X}_{\left( n \right)}}} \right)} \right)$, where ${\mathbf{X}_{\left( i \right)}}$ denotes the mode-$i$ unfolding matrix of ${\cal X}$. Generalized from the nuclear norm in matrix case, Liu et al. \cite{b1} first defined the sum of nuclear norm (SNN) as a convex relaxation of the tensor n-rank:
\begin{equation}
{\left\| {{\cal X}} \right\|_{SNN}} = \sum\limits_{i = 1}^n {{{\left\| {{\mathbf{X}_{\left( i \right)}}} \right\|}_*}}.\label{eq4}
\end{equation}

where ${\left\|  \cdot  \right\|_*}$ denotes the nuclear norm of a matrix. In addition to the basic definition of SNN, an average version was proposed by Gandy et al. \cite{b22} for tensor completion. In particular, Signoretto et al. \cite{b23} further generalized SNN to the Shatten-$p,q$ norm. Considering the subspace structure in each mode, Kasai et al. \cite{b24} developed a LRTC method based on the Riemannian manifold. Moreover, many other norms based on SNN such as latent trace norm and scaled latent trace norm were proposed to be an approximation of the tensor n-rank. More recently, combining the tensor total variation and SNN, Yokota et al. \cite{b25} proposed a new LRTC model that can be solved based on primal-dual splitting framework.

However, several limitations of SNN occur with the increasing tensor dimension and scale, including: 1) the operation that simply unfold tensor into matrices along each mode ignores the tensor’s intrinsic structure. 2) The SNN model is neither the tightest convex lower bound of tensor n-rank nor the optimal solution with the dimension increasing. 3) The unfolding and folding are expensive.

Unlike the unfolding method of tensor n-rank, the tensor train (TT) rank \cite{b26} employs a well-balanced matricization scheme to capture the global correlation of tensor entries. For a given tensor ${{\cal X}} \in {^{{I_1} \times  \cdots  \times {I_n}}}$, Imaizumi et al. \cite{b27} and Bengua et al. \cite{b26} matricize the tensor along permutations of modes and define the tensor train nuclear norm (TTNN) based on TT rank as:
\begin{equation}
{\left\| {{\cal X}} \right\|_{TTNN}} = \sum\limits_{i = 1}^{n - 1} {{{\left\| {{\mathbf{X}_{\left[ i \right]}}} \right\|}_*}}.\label{eq5} 
\end{equation}
where ${X_{\left[ i \right]}} \in {^{{m_1} \times {m_2}}}$, ${m_1} = \prod\nolimits_{k = 1}^i {{I_k}} $ and ${m_2} = \prod\nolimits_{k = i + 1}^n {{I_k}} $ denotes the mode-$\left( {1, \cdots ,i} \right)$ unfolding matrix of ${\cal X}$. Imaizumi et al. \cite{b27} established the statistical theory and developed a scalable algorithm for the TTNN model. They provided a statistical error bound for TTNN which achieves the same efficiency as SNN. Moreover, Bengua et al. \cite{b26} stated that tensor train nuclear norm is more tractable than SNN for LRTC, and they introduced ket augmentation scheme to obtain a higher order tensor from a given tensor. Furthermore, algorithmic development for the solution of TTNN model was also devoted in the research.

Although the matricization scheme ${\mathbf{X}_{\left[ i \right]}}$ is more balance than the unfolding ${\mathbf{X}_{\left( i \right)}}$ , the matricize operation may also destroy the original tensor’s internal structure. Moreover, the TTNN model is efficient for higher order tensors.

To avoid the tensor matricization and maintain the intrinsic data structure, Kilmer et al. \cite{b17} proposed tensor singular value decomposition (t-SVD) and this motivates the new tensor multi-rank and tubal rank. Another advantage of such method is that the resultant algebra and analysis are very close to those of matrix case. Zhang et al. \cite{b28} give the definition of a new tensor nuclear norm (TNN) corresponding to tubal rank and t-SVD. Leveraging the conclusion of matrix case, they state that TNN is the tightest convex approximation of tensor average rank. Furthermore, they derived the exact recovery conditions for LRTC problems in \cite{b29}.

Due to the advantages of TNN, it has received more and more attention in recent years. Nevertheless, the computational complexity of TNN increases dramatically for the large scale tensor. Additionally, TNN penalizes all the singular values with the same weight. In fact, the larger the singular value, the more information it contains, and the less penalized it should be. To conquer the challenge of high complexity, Zhou et al. \cite{b30} focus on utilizing the technique of tensor factorization for LRTC problems. Only two smaller tensors are maintained in the optimization process, which can be used to preserve the low-rank structure of the underlying tensor. In addition, extending Schatten-$p$ norm to tensor space, Kong et al. \cite{b31} state that their propose tensor Schatten-$p$ norm can better approximate the tensor average rank.

Recently, Voronin et al. \cite{b32} generalize the iterative soft thresholding method to $p$-shrinkage thresholding for solving sparse signal recovery (SSR) problems. Both theoretical and empirical results in SSR problem prove that the $p$-shrinkage thresholding function is a good alternative to the Schatten-$p$ norm, which can achieve better recovery performance than the existing surrogates. Motivated by this success application, we attempt to introduce the $p$-shrinkage scheme to LRTC problem and develop an efficient algorithm to solve the resulting model.

In this paper, we propose a new tensor $p$-shrinkage nuclear norm ($p$-TNN), which is defined in \eqref{eq13} based on the t-SVD and $p$-shrinkage scheme. When $ - \infty  < p < 1$, the proposed norm can achieve a better approximation of the tensor average rank than TNN, i.e., $p$-TNN is a tighter surrogate. Extending the matrix norm surrogate to the tensor case, we establish the $p$-TNN based LRTC model. Additionally, the recovery guarantee of the proposed model is provided. To cope with the challenges of the resultant non-convex optimization problem, we develop an efficient algorithm under the alternating direction method of multipliers (ADMM) framework. Furthermore, we also incorporate the adaptive momentum scheme to accelerate the empirical convergence for the proposed algorithm. Subsequently, the resulting algorithm is analyzed in detail from the aspects of time complexity and convergence, respectively.

In summary, the primary contributions of our work include:

\begin{itemize}
	\item \emph{We propose a new definition of tensor $p$-shrinkage nuclear norm with some desirable properties, for example, positivity and unitary invariance. The proposed $p$-TNN ($ - \infty  < p < 1$) is a tighter envelope of the tensor average rank than TNN within the unit ball of the spectral norm, which is beneficial to improve the recovery performance of LRTC problem.}
	
	\item \emph{By employing the tensor $p$-shrinkage nuclear norm, we propose a novel LRTC model and provide a strong guarantee for tensor recovery, i.e., the error in recovering a ${I_1} \times {I_2} \times {I_3}$ tensor is 
	${\cal O}\left({{{r{I_1}{I_3}\log \left( {{3 \mathord{\left/{\vphantom {3 \alpha }} \right.\kern-\nulldelimiterspace} \alpha }} \right)}}/{{\left| \Omega \right|}}}\right)$, where $\alpha  = {\textstyle{3 \over {\left( {{I_1} + {I_2}} \right){I_3}}}}$, $r$ denotes the tensor tubal rank, and $\left| \Omega  \right|$ denotes the cardinal number of the index set.}
	
	\item \emph{Incorporating the adaptive momentum scheme, we develop an efficient algorithm which establishes a unified framework for algorithms that applying the soft and hard thresholding shrinkage. Under the smoothness assumption, the Convergence guarantee to critical points is provided.}
\end{itemize}

\section{Notations and preliminaries}
In this section, we first introduce some main basic notations and then briefly give some necessary definitions which will be used later.

Tensors are denoted by uppercase calligraphy letters, e.g., ${\cal {A}}$. Matrices are denoted by uppercase boldface letters, e.g., $\mathbf{A}$. Vectors are denoted by lowercase boldface letters, e.g., $\mathbf{a}$, while scalars are denoted by lowercase letters, e.g., $a$.

The fields of complex numbers and real numbers are denoted by ${\mathbb {C}}$ and ${\mathbb {R}}$, respectively. We simply represent ${\left\{1,2, \cdots n\right\}}$ by ${\left[n\right]}$. For a n-dimensional tensor ${{\cal {A}} \in \mathbb{C}{^{{I_1} \times {I_2} \times \cdots \times {I_n}}}}$, we use ${\cal A}_{i_1i_2 \cdots i_n}$ to represent its $\left({i_1i_2 \cdots i_n}\right)$-th entry, where ${i_k \in \left[I_k\right]}$ and ${k \in {\left[n\right]}}$. Let ${\mathbf{A}^{\left(k\right)}}$ denote the $k$-th frontal slice of the tensor ${\cal {A}}$. We use ${\cal A}^{\top}$ to denote its transpose tensor. The inner product of ${\cal {A}}$ and ${\cal {B}}$ is defined as $\left\langle {{\cal A},{\cal B}} \right\rangle = \sum\nolimits_{{i_1},{i_2}, \cdots ,{i_n}} {{{\cal A}_{{i_1},{i_2}, \cdots ,{i_n}}}} {{\cal B}_{{i_1},{i_2}, \cdots ,{i_n}}}$, and the Frobenius norm of ${\cal A}$ is defined as ${{\left\| {\cal A} \right\|_F} = \sqrt {\left\langle {{\cal A},{\cal A}} \right\rangle }  = \sqrt {\sum\nolimits_{{i_1},{i_2}, \cdots ,{i_n}} {\left(A_{{i_1},{i_2}, \cdots ,{i_n}}\right)^2} }}$. 

For a given 3-dimensional tensor ${\cal X} \in {\mathbb{R}^{I_1 \times I_2 \times I_3}}$, we use ${\bar {\cal X}}\in {\mathbb{C}^{I_1 \times I_2 \times I_3}}$ to denote the Discrete Fourier Transformation (DFT) of ${\cal X}$ along the third dimension, i.e., ${\bar{\cal X}} = fft{\left({\cal X},[{\kern 1pt} {\kern 1pt}],3\right)}$. Correspondingly, the $k$-th frontal slice of ${\bar{\cal X}}$ is denoted by ${{\bar {\mathbf{X}}}^{\left(k\right)}}$. The unfold and its inverse operator of $\cal X$ \cite{b2} are defined as
\begin{equation}
unfold\left( {\cal X} \right) = \left[ {\begin{array}{*{20}{c}}
	{{{\mathbf X}^{\left( 1 \right)}}}\\
	{{{\mathbf X}^{\left( 2 \right)}}}\\
	\vdots \\
	{{{\mathbf X}^{\left( {{I_3}} \right)}}}
	\end{array}} \right], fold\left( {unfold\left( {\cal X} \right)} \right) = {\cal X}.\label{eq6} 
\end{equation}

And the block circulant matrix of ${\cal X}$ is further defined as ${bcirc \left({\cal X}\right)} \in \mathbb{R}^{I_1I_3 \times I_2I_3}$ \cite{b2}:
\begin{equation}
bcirc \left({\cal X}\right) = \left[ {\begin{array}{*{20}{c}}
{{\mathbf{X}^{\left( 1 \right)}}}&{{\mathbf{X}^{\left( I_3 \right)}}}& \cdots &{{\mathbf{X}^{\left( 2 \right)}}}\\
{{\mathbf{X}^{\left( 2 \right)}}}&{{\mathbf{X}^{\left( 1 \right)}}}& \cdots &{{\mathbf{X}^{\left( 3 \right)}}}\\
\vdots & \vdots & \ddots & \vdots \\
{{\mathbf{X}^{\left( I_3 \right)}}}&{{\mathbf{X}^{\left( I_3-1 \right)}}}& \cdots &{{\mathbf{X}^{\left( 1 \right)}}}
\end{array}} \right].\label{eq7} 
\end{equation}

Then we summarize some necessary definitions and results.\\
\textbf{Definition 1. (t-product \cite{b17})} For two given tensors ${\cal A} \in {\mathbb{R}^{I_1 \times d \times I_3}}$ and ${\cal B} \in {\mathbb{R}^{d \times I_2 \times I_3}}$, the t-product of $\cal A$ and $\cal B$ is defined as:
\begin{equation}
{\cal C} = {\cal A} * {\cal B} = fold\left( {bcirc\left( {\cal A} \right) \cdot unfold\left( {\cal B} \right)} \right).\label{eq8}
\end{equation}
where ${\cal C} \in {\mathbb{R}^{I_1 \times I_2 \times I_3}}$. Therefore, the t-product is homologous in form to the matrix multiplication except that the operation of circulant convolution and unfolding. Additionally, we can also note that if $I_3 = 1$, the t-product reduces to matrix multiplication.

In addition, the definitions of orthogonal tensor, identity tensor, frontal-slice-diagonal tensor (f-diagonal tensor)and tensor transpose can be found in the Appendix. Subsequently, the Tensor Singular Value Decomposition (t-SVD) can be defined as follows by using these above definitions.\\
\textbf{Definition 2.} (t-SVD \cite{b17}) For a given tensor ${\cal X} \in {\mathbb{R}^{I_1 \times I_2 \times I_3}}$, there exist ${\cal U} \in {\mathbb{C}^{I_1 \times I_1 \times I_3}}$, ${\cal V} \in {\mathbb{C}^{I_2 \times I_2 \times I_3}}$ and ${\cal S} \in {\mathbb{R}^{I_1 \times I_2 \times I_3}}$ such that:
\begin{equation}
{\cal X} = {\cal U} * {\cal S} * {{\cal V}^{\top}}.\label{eq9}
\end{equation}
where ${\cal U}$ and ${\cal V}$ are orthogonal tensor, i.e., ${\cal U}*{\cal U}^{\top} = {\cal V}*{\cal V}^{\top} = {\cal I}$, and ${\cal S}$ is a f-diagonal tensor.
\\
\textbf{Definition 3.} (Tensor tubal rank \cite{b28} and multi-rank \cite{b29}) For a given tensor ${\cal X} \in {\mathbb{R}^{I_1 \times I_2 \times I_3}}$, then the tensor tubal rank of ${\cal X}$, herein denoted by $rank_t \left({\cal X}\right)$, is defined to be the number of nonzero singular tubes of ${\cal S}$, i.e., 
\begin{equation}
rank_t\left({\cal X}\right) = \sum\nolimits_{k = 1}^{\min \left\{ {{I_1},{I_2}} \right\}} {{\mathbf{I}_{{\cal S}\left( {k,k,:} \right) \ne 0}}}.\label{eq10}
\end{equation}
where ${\cal S}$ is defined in Definition 2, ${\cal S}\left( {k,k,:} \right)$ denotes the $k$-th diagonal tube of ${\cal S}$, and $\mathbf{I}_{{\cal S}\left( {k,k,:} \right) \ne 0}$ is an indicator function, i.e., $\mathbf{I}_{{\cal S}\left( {k,k,:} \right) \ne \mathbf{0}} = 1$ if ${{\cal S}\left( {k,k,:} \right) \ne \mathbf{0}}$ is true and $\mathbf{I}_{{\cal S}\left( {k,k,:} \right) \ne \mathbf{0}} = 0$ otherwise. The tensor multi-rank of ${\cal X}$ is a vector $\mathbf{r} \in {\mathbb{R}^{I_3}}$ consisting of the rank of the frontal slice of $\bar{\cal X}$, i.e., ${\mathbf r}_k = rank\left( {{{\bar {\mathbf{X}}}^{\left( k \right)}}} \right)$.\\
\textbf{Definition 4.} (Tensor nuclear norm and spectral norm \cite{b33}) For a given tensor ${\cal X} \in {\mathbb{R}^{I_1 \times I_2 \times I_3}}$, the tensor nuclear norm of ${\cal X}$, denoted by ${\left\| {\cal X} \right\|_*}$, is defined as the average of the nuclear norm of all the frontal slices of ${\bar {\cal X}}$:
\begin{equation}
{\left\| {\cal X} \right\|_*} = {\frac{1}{I_3}}{\sum\limits_{k = 1}^{I_3} {\left\| {{{\bar X}^{\left( {k} \right)}}} \right\|} _*}.\label{eq11}
\end{equation}
Furthermore, the tensor spectral norm of $\cal X$, denoted by ${\left\| {{\bar {\cal X}}} \right\|}$, is defined as ${\left\| {{\bar {\cal X}}} \right\|} = \mathop {\max }\limits_i \left\| {{{\bar {\mathbf{X}}}^{\left( k \right)}}} \right\|$.

According to the Von Neumann's inequality, it can be seen that the tensor spectral norm is the dual norm of the tensor nuclear norm and vice versa.
\\
\textbf{Definition 5.} ($p$-shrinkage thresholding operator \cite{b32}) For a given scalar $x \in {\mathbb{R}}$, $\forall \mu  > 0$ and $p \le 1$, the $p$-shrinkage thresholding operator, denoted by $s_p^\mu$, is defined as:
\begin{equation}
{s_p^\mu \left(x\right)} = sign\left(x\right)max{\left\{\left| x \right| - \mu {\left| x \right|^{p - 1}},0\right\}}.\label{eq12}
\end{equation}
where $sign\left(x\right)$ denotes the sign function. 
\begin{figure}[h]
	\centering
	\includegraphics[width=0.5\textwidth]{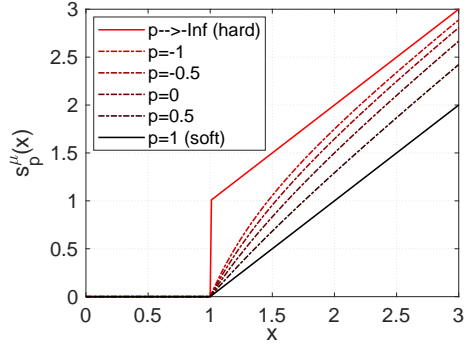}
	\caption{Several $p$-shrinkage functions with different $p$ values. Without loss of generality, the $\mu$ is fixed at 1. The smaller the $p$ value, the less penalty for large inputs.}
	\label{Fig.1}
\end{figure}

Several $p$-shrinkage functions with different $p$ values are shown in Fig.\ref{Fig.1}. Note that if $p = 1$, the $p$-shrinkage thresholding transforms to the soft thresholding, which imposes no different penalty for all the inputs and is commonly used in matrix case, and if $p \to -\infty$, it turns into the hard thresholding with no penalty for large inputs. However, hard thresholding is discontinuous. Moreover, when $- \infty < p < 1$, the $p$-shrinkage function satisfies the property that the larger the input, the less penalized it should be.

\section{Model}
In this section, we propose a new definition of tensor $p$-shrinkage nuclear norm ($p$-TNN) based on $p$-shrinkage scheme and t-SVD. Subsequently, we propose a novel LRTC model by employing our $p$-TNN.\\
\textbf{Definition 5.} (Tensor $p$-shrinkage nuclear norm, $p$-TNN) For a given tensor ${\cal X} \in {\mathbb{R}^{I_1 \times I_2 \times I_3}}$, let ${\cal X}={\cal U} * {\cal S} * {\cal V}^{\top}$ be the tensor singular value decomposition of ${\cal X}$. Then the tensor $p$-shrinkage nuclear norm is defined as:
\begin{equation}
{\left\| {\cal X} \right\|_p} := \frac{1}{{I_3}}{\sum\limits_{k = 1}^{I_3} {\left\| {{{\bar {\mathbf{X}}}^{\left( k \right)}}} \right\|} _p} := \frac{1}{{{I_3}}}\sum\limits_{k = 1}^{{I_3}} {\sum\limits_{i=1}^{\min \left\{ {{I_1},{I_2}} \right\}} {S_p^\mu\left( {{\bar {\cal S}}} \right)} }.\label{eq13}
\end{equation}
where $-\infty<p<1$, and ${S_p^\mu\left( {{\bar {\cal S}}} \right)}: {\mathbb{R}^{I_1 \times I_2 \times I_3}} \to {\mathbb{R}^{I_1 \times I_2 \times I_3}}$ is defined as follows:
\begin{equation}
\left({S_p^\mu\left( {{\bar {\cal S}}} \right)}\right)_{iik} 
= {s_p^\mu\left( {{\bar {\cal S}_{iik}}} \right)}
=max\left\{\left|{\bar {\cal S}_{iik}}\right|-\mu{\left|{\bar {\cal S}_{iik}}\right|^{p-1}} ,0 \right\}.\label{eq14}
\end{equation}
where ${{\bar {\cal S}_{kki}}}$ denotes the $\left(k,k,i\right)$-th tensor singular value of ${\bar {\cal X}}$.

The following are some of the properties of our proposed $p$-TNN that we use in this paper. For the proofs, please refer to the Appendix.
\\\\
\textbf{Proposition 1. (Positivity)} For a given tensor ${\cal X} \in {\mathbb{R}^{I_1 \times I_2 \times I_3}}$, the $p$-TNN of ${\cal X}$, 
denoted by ${\left\| {\cal{X}} \right\|_p}$. Obviously, ${\left\| {\cal{X}} \right\|_p} \ge 0$ with equality holding if and only if ${\cal X}$ is zero. Moreover, ${\left\| {\cal{X}} \right\|_p}$ is non-decreasing.
\\\\
\textbf{Proposition 2. (Non-convexness)} For a given tensor ${\cal X} \in {\mathbb{R}^{I_1 \times I_2 \times I_3}}$, if $-\infty<p<1$, ${\left\| {\cal{X}} \right\|_p}$ is non-convex w.r.t. ${\cal X}$. i.e., ${\left\| {\cal{X}} \right\|_p}$ cannot satisfy the inequality for any $\rho \in \left(0,1\right)$:
\begin{equation}
{\left\| {\rho {{{\cal X}}_1} + \left( {1 - \rho } \right){{{\cal X}}_2}} \right\|_p} \le \rho {\left\| {{{{\cal X}}_1}} \right\|_p} + \left( {1 - \rho } \right){\left\| {{{{\cal X}}_2}} \right\|_p}.\label{eq15}
\end{equation}
where ${\cal X}_1,{\cal X}_2 \in {\mathbb{R}^{I_1 \times I_2 \times I_3}}$ and ${\cal X}_1 \ne {\cal X}_2$.
\\\\
\textbf{Proposition 3. (Unitary invariance)} For a given tensor ${\cal X} \in {\mathbb{R}^{I_1 \times I_2 \times I_3}}$, there exist orthogonal tensors ${\cal U} \in {\mathbb{R}^{I_1 \times I_2 \times I_3}}$ and ${\cal V} \in {\mathbb{R}^{I_1 \times I_2 \times I_3}}$, such that:
\begin{equation}
{\left\| {{\cal X}} \right\|_p} = 
{\left\| {{{\cal U}}*{{\cal X}}*{{{\cal V}}^{\top}}} \right\|_p}.\label{eq16}
\end{equation}
In addition, ${\left\| {{\cal X}} \right\|_p} = {\left\| {{{\cal U}}*{{\cal X}}} \right\|_p} = {\left\| {{{\cal X}}*{{{\cal V}}^{\top}}} \right\|_p}$ also holds. Therefore, the tensor $p$-shrinkage unclear norm is unitary invariance.
\\

Extending the proximal operator to the tensor case, then for any $\tau > 0$, we can define the proximal operator of the tensor $p$-shrinkage nuclear norm ${\left\| { \cdot  } \right\|_p}$ as:
\begin{equation}
pro{x_{{\tau }{{\left\| {{\cal X}} \right\|}_p}}}\left( {{\cal Z}} \right) = \arg \mathop {\min }\limits_{{\cal X}} \frac{1}{2}\left\| {{{\cal X}} - {{\cal Z}}} \right\|_F^2 + {\tau }{\left\| {{\cal X}} \right\|_p}.\label{eq17}
\end{equation}

The following definition extends the generalized singular value thresholding (GSVT) to the tensor space. Similar to the matrix case, the following shows that the proximal operator of the tensor $p$-shrinkage nuclear norm has a closed-form solution.
\\\\
\textbf{Definition 6.} (Tensor generalized singular value thresholding, t-GSVT) Let ${\cal X},{\cal Z} \in {\mathbb{R}^{I_1 \times I_2 \times I_3}}$, for any $\tau > 0$, the solution of the proximal operator of $p$-TNN can be calculated as:
\begin{equation}
pro{x_{{\tau }{{\left\| {{\cal X}} \right\|}_p}}}\left( {{\cal Z}} \right) = {\cal U} * {\cal D} * {\cal V}^{\top}.\label{eq18}
\end{equation}
where ${\cal U} * {\cal S} * {\cal V}^{\top}$ is the t-SVD of ${\cal Z}$, and the f-diagonal tensor ${\cal D} \in {\mathbb{R}^{I_1 \times I_2 \times I_3}}$ that satisfies ${\bar{\cal D}} = S_p^\mu \left({\bar{\cal S}}\right)$.
\\

For a given tensor ${\cal X} \in {\mathbb{R}^{I_1 \times I_2 \times I_3}}$, Lu et al. \cite{b33} defined the tensor average rank as $rank_a\left({\cal X}\right) = \frac{1}{{I_3}}\sum\nolimits_{k = 1}^{I_3} {rank\left( {{{\bar {\mathbf{X}}}^{\left( i \right)}}} \right)}$, and they pointed out that TNN is the convex envelope of the tensor average rank within the unit ball of the tensor spectral norm. Recently, Kong et al. \cite{b31} proposed a tensor Schatten-$p$ norm and claimed that it can be a tighter non-convex approximation of the tensor average rank than TNN.  Here, we prove that the proposed $p$-TNN ${\left\| {\cal X} \right\|_p}$ is even a tighter envelope within the same unit ball.
\\\\
\textbf{Proposition 4. (Tightness)} For a given tensor ${\cal X} \in {\mathbb{R}^{I_1 \times I_2 \times I_3}}$, when $-\infty < p < 1$, ${\left\| {\cal X} \right\|_p}$ is an non-convex envelope of the tensor average rank within the unit ball of the spectral norm, which is tighter than TNN in the sense of $\left\| {{\cal X}} \right\|_* \le {\left\| {{\cal X}} \right\|_p} \le rank_a\left({\cal X}\right)$.
\\

Considering the LRTC model defined in \eqref{eq2}, the non-smoothness rank function is usually relaxed as its convex/non-convex surrogate. By using Definition 4 and its Proposition 4, we can rewrite the tensor completion problem into the following form:
\begin{equation}
\mathop {\min }\limits_{\cal X} \frac{1}{2}\left\| {{{\cal P}_\Omega }\left( {{\cal X} - {\cal T}} \right)} \right\|_F^2 + \lambda {\left\| {\cal X} \right\|_p}.\label{eq19}
\end{equation}
where ${\cal T} \in {\mathbb{R}^{I_1 \times I_2 \times I_3}}$ denotes the observed tensor, $\Omega$ denotes the index set of the observed entries, and ${\cal P}_\Omega$ denotes the projection operator, i.e., ${\left[ {{{\cal P}_\Omega }\left( {\cal A} \right)} \right]_{i_1i_2i_3}} = {{\cal A}_{i_1i_2i_3}}$ if $\left(i_1,i_2,i_3\right) \in \Omega$ and 0 otherwise. We aim to recover the completed tensor (underlying tensor) ${\cal X}$ based on the observed missing tensor ${\cal T}$.
\\
\textbf{Recovery guarantee.} Following the definition of the recovery error in \cite{b34}, here the the recovery error can be formulated as:
\begin{equation}
{\rm{R}}\left({\cal X}\right) = \frac{1}{I_1I_2I_3}\left\| {\cal X} - {\cal X^{*}}\right\|_F^2.\label{eq20}
\end{equation}
where ${\cal X},{\cal X}^* \in {\mathbb{R}^{I_1 \times I_2 \times I_3}}$ are the underlying tensor and its recovery tensor, respectively. The following theorem establishes the upper bound of the recover error based on our proposed $p$-TNN.
\\\\
\textbf{Theorem 1.} (Main Result 1) Let ${\cal X}^* \in {\mathbb{R}^{I_1 \times I_2 \times I_3}}$ be the solution to \eqref{eq19}, then with a probability of at least$1-\alpha$, we have
\begin{equation}
\frac{1}{I_1I_2I_3}\left\| {\cal X} - {\cal X^{*}}\right\|_F^2 \le
Cmax\left\{\frac{{{r}I_1I_3\log \left({3 \mathord{\left/{\vphantom {3 \alpha }} \right.
\kern-\nulldelimiterspace} \alpha }\right)}}{{\left| \Omega  \right|}},\sqrt {\frac{{\log \left( {{3 \mathord{\left/{\vphantom {3 \alpha }} \right.\kern-\nulldelimiterspace} \alpha }} \right)}}{{\left| \Omega  \right|}}} \right\}.\label{eq21}
\end{equation}
where $C$ is an absolute constant, $r$ denotes the tensor tubal rank of the underlying tensor, ${\left| \Omega  \right|}$ denotes the cardinal number of the index set $\Omega$, and $\alpha = {\textstyle{3 \over {\left( {I_1 + I_2} \right)I_3}}}$. The proof can be found in Appendix.
\\

According to \eqref{eq21}, it can be seen that the upper bound of the recovery error is related to the tubal of rank the underlying tensor, as well as the sampling rate $sr$, which can be calculated by $sr = {\textstyle{{\left| \Omega  \right|} \over {I_1I_2I_3}}}$.

When using a tighter surrogate, one can get a better solution. From the above narrative, we can see that the main advantage of $p$-TNN is the outstanding tightness (Proposition 4), i.e., our $p$-TNN is a better approximation of the tensor average rank, which can lead to a better solution. However, $\left\| {\cal X} \right\|_p$ is non-convex when $-\infty < p < 1$. The resulting non-convex optimization problem is much more challenging. A strong performance guarantee would be hard to get as in the convex case.

\section{Algorithm}
In this section, we will show that the proposed non-convex model \eqref{eq19} can be solved effectively based on the ADMM framework. Moreover, we will show that the proposed algorithm can be further accelerated with adaptive momentum.

Let $F\left( {{\cal X}} \right) \equiv \frac{1}{2} \left\| {{{{\cal P}}_\Omega }\left( {{{\cal X}} - {{\cal T}}} \right)} \right\|_F^2 + \lambda {\left\| {{\cal X}} \right\|_p}$. By introducing an auxiliary variable ${\cal Y} \in {\mathbb{R}^{I_1 \times I_2 \times I_3}}$, such that ${\cal Y} = {\cal X}$, the augmented Lagrangian function of \eqref{eq19} is given as follows:
\begin{equation}
{{\cal L}}\left( {{{\cal X}},{{\cal Y}},{{\cal Z}},\beta } \right) = \lambda {\left\| {{\cal Y}} \right\|_p} + \frac{1}{2}\left\| {{{{\cal P}}_\Omega }\left( {{{\cal X}} - {{\cal T}}} \right)} \right\|_F^2 + \left\langle {{{\cal Z}},{{\cal X}} - {{\cal Y}}} \right\rangle  + \frac{\beta }{2}\left\| {{{\cal X}} - {{\cal Y}}} \right\|_F^2.\label{eq22}
\end{equation}
where ${\cal Z} \in {\mathbb{R}^{I_1 \times I_2 \times I_3}}$ denotes the Lagrangian multiplier, and $\beta$ denotes the penalty parameter. In the following, \eqref{eq22} can be solved by applying the classical ADMM framework. Therefore, ${\cal X}$, ${\cal Y}$, ${\cal Z}$ and $\beta$ are iteratively update as:
\begin{equation}
\left\{ \begin{aligned}
{{{\cal Y}}^{t + 1}} &= \arg \mathop {\min }\limits_{{\cal Y}} {{\cal L}}\left( {{{{\cal X}}^t},{{\cal Y}},{{{\cal Z}}^t},{\beta ^t}} \right)\\
{{{\cal X}}^{t + 1}} &= \arg \mathop {\min }\limits_{{\cal X}} {{\cal L}}\left( {{{\cal X}},{{{\cal Y}}^{t + 1}},{{{\cal Z}}^t},{\beta ^t}} \right)\\
{{{\cal Z}}^{t + 1}} &= {{{\cal Z}}^t} + {\beta ^t}\left( {{{{\cal Y}}^{t + 1}} - {{{\cal X}}^{t + 1}}} \right)\\
{{\beta}^{t + 1}}    &= \min \left( {\eta {\beta ^t},{\beta ^{\max }}} \right)
\end{aligned} \right.\label{eq23}
\end{equation}
\\
\textbf{1) For ${\cal Y}$-subproblem}.

By fixing ${\cal X}^t$, ${\cal Z}^t$ and $\beta^t$, we can obtain ${\cal Y}^{t+1}$ by:
\begin{equation}
\begin{aligned}
{{{\cal Y}}^{t + 1}} 
&= \arg \mathop {\min }\limits_{{\cal Y}} \lambda {\left\| {{\cal Y}} \right\|_p} + \left\langle {{{{\cal Z}}^t},{{{\cal X}}^t} - {{\cal Y}}} \right\rangle  + \frac{\beta }{2}\left\| {{{{\cal X}}^t} - {{\cal Y}}} \right\|_F^2\\
&= \arg \mathop {\min }\limits_{{\cal Y}} \lambda {\left\| {{\cal Y}} \right\|_p} + \frac{\beta }{2}\left\| {{{{\cal X}}^t} - {{\cal Y}} + \frac{1}{\beta }{{{\cal Z}}^t}} \right\|_F^2
\end{aligned}.\label{eq24}
\end{equation}

It can be seen that \eqref{eq24} satisfies the form of the proximal operator of $p$-TNN. Therefore, according to \eqref{eq17}, ${\cal Y}^{t+1}$ can be rewritten as:
\begin{equation}
{{{\cal Y}}^{t + 1}} = pro{x_{\lambda {{\left\| {{\cal Y}} \right\|}_p}}}\left( {{{{\cal X}}^t} - \frac{1}{\beta }{{{\cal Z}}^t}} \right).\label{eq25}
\end{equation}

\textbf{2) For ${\cal X}$-subproblem}.

By fixing ${\cal Y}^{t+1}$, ${\cal Z}^t$ and $\beta^t$, we can obtain ${\cal X}^{t+1}$ by:
\begin{equation}
\begin{aligned}
{{{\cal X}}^{t + 1}} &= \arg \mathop {\min }\limits_{{\cal X}} \frac{1}{2}\left\| {{{{\cal P}}_\Omega }\left( {{{\cal X}} - {{\cal O}}} \right)} \right\|_F^2 + \left\langle {{{{\cal Z}}^t},{{\cal X}} - {{{\cal Y}}^{t + 1}}} \right\rangle  + \frac{\beta }{2}\left\| {{{\cal X}} - {{{\cal Y}}^{t + 1}}} \right\|_F^2\\
&= \arg \mathop {\min }\limits_{{\cal X}} \frac{1}{2}\left\| {{{{\cal P}}_\Omega }\left( {{{\cal X}} - {{\cal O}}} \right)} \right\|_F^2 + \frac{\beta }{2}\left\| {{{\cal X}} - {{{\cal Y}}^{t + 1}} + \frac{1}{\beta }{{{\cal Z}}^t}} \right\|_F^2
\end{aligned}.\label{eq26}
\end{equation}
The above equation has a closed-form solution as following:
\begin{equation}
{{{\cal X}}^{t + 1}} = {\cal P}_{\bar{\Omega}}{\left( {{{{\cal Y}}^{t + 1}} + \frac{1}{\beta }{{{\cal Z}}^t}} \right)} + {{\cal T}}.\label{eq27}
\end{equation}
where ${\bar{\Omega}}$ denotes the complementary set of the index set $\Omega$.

Next, we will show the above update step can be accelerated by the adaptive momentum, which has been popularly used for stochastic gradient descent and proximal algorithms. The idea of the adaptive momentum is to apply historical iterations to accelerate convergence. The whole procedure is shown in Algorithm 1. Let the underlying tensor ${\cal X} \in {\mathbb{R}^{I_1 \times I_2 \times I_3}}$, we suppose that the tubal rank of $\cal X$ is $rank_t\left({\cal X}\right) = r$. Obviously, the main per-iteration cost lies in the update of ${\cal Y}^{t+1}$ (step 9 in Algorithm 1), which takes ${\cal O}\left(r\left(I_1+I_2\right)I_3logI_3+r\left|\Omega\right|\right)$ time. As for TNN in \cite{b33}, the computational complexity at each iteration is ${\cal O}\left(I_1I_2I_3\left(logI_3 + min\left\{I_1,I_2\right\}\right)\right)$. With $r < min\left\{I_1,I_2\right\}$ and $\left|\Omega\right| < I_1I_2I_3$, it can be seen that our algorithm is much more efficient than TNN based methods in each iteration.
\begin{algorithm}
	\caption{ADMM for solving \eqref{eq19}.}
	\LinesNumbered
	\KwIn{the observed tensor ${\cal O}$, the observed index set $\Omega$, and parameters $p < 1$, $\rho > 1$, $\gamma$, $\lambda$, $\beta^0$, and $\beta^{max}$, the maximum number of iterations $T$.}
	initialize ${\cal X}^{-1} = {\cal X}^{0} = \mathbf{0}$, ${\cal Y}^0 = {\cal Z}^0 = {\mathbf{0}}$, $\eta$ = 1.1;\\
	\While{not converged}{
		${\cal Q}^t = {\cal X}^t + \gamma^t\left({\cal X}^t-{\cal X}^{t-1}\right)$;\\
		\If {$F\left({\cal Q}^t\right) \le F\left({\cal X}^t\right)$}{${\cal W}^t = {\cal Q}^t, \gamma^{t+1} = min\left\{1,\rho\gamma^t\right\};$\\
		\textbf{else}\\
		${\cal W}^t = {\cal X}^t, \gamma^{t+1} ={\gamma^t  \mathord{\left/
				{\vphantom {\gamma^t  \rho }} \right.
				\kern-\nulldelimiterspace} \rho };$
		}
		Update ${\cal Y}^{t+1}$ by
		$${{{\cal Y}}^{t + 1}} = pro{x_{\lambda {{\left\| {{\cal Y}} \right\|}_p}}}\left( {{{{\cal W}}^t} - \frac{1}{\beta ^t}{{{\cal Z}}^t}} \right);$$\\
		Update ${\cal X}^{t+1}$ by
		$${{{\cal X}}^{t + 1}} = {\cal P}_{\bar{\Omega}}{\left( {{{{\cal Y}}^{t + 1}} + \frac{1}{\beta^t }{{{\cal Z}}^t}} \right)} + {{\cal T}};$$\\
		Update ${\cal Z}^{t+1}$ by
		$${{{\cal Z}}^{t + 1}} = {{{\cal Z}}^t} + {\beta ^t}\left( {{{{\cal Y}}^{t + 1}} - {{{\cal X}}^{t + 1}}} \right);$$\\
		Update ${\beta}^{t + 1}$ by
		${{\beta}^{t + 1}}    = \min \left( {\eta {\beta ^t},{\beta ^{\max }}} \right);$\\
		Check the convergence conditions
		$$t >　T, {{{{\left\| {{{\cal X}^{t + 1}} - {{\cal X}^t}} \right\|}_F}} \mathord{\left/
				{\vphantom {{{{\left\| {{{\cal X}^{t + 1}} - {{\cal X}X^t}} \right\|}_F}} {{{\left\| {{{\cal X}^t}} \right\|}_F}}}} \right.
				\kern-\nulldelimiterspace} {{{\left\| {{{\cal X}^t}} \right\|}_F}}} \le tol.$$\\
		$t \leftarrow t+1.$
	}
	\KwResult {The recovery tensor {${{{\cal X}}^*}$}.}
\end{algorithm}
\\
\textbf{Convergence analysis.}
In this section, we investigate the convergence of the Algorithm 1. $\left({\cal X}^*,{\cal Y}^*,{\cal Z}^*\right)$ is the KKT point of the problem (14), if it satisfies the following system:
\begin{equation}
\left\{ \begin{aligned}
	{{{\cal Y}}^ * } &= {{{\cal U}}^ * }*{{{\cal D}}^ * }*{{{\cal V}}^ * }^\top\\
	{{{\cal X}}^ * } &= {{{\cal Y}}^ * }\\
	\mathbf 0 &=\frac{1}{\beta }\left( {{{{\cal X}}^ * } - {{\cal T}}} \right) + {{{\cal Z}}^ * }
\end{aligned} \right.\label{eq28}
\end{equation}
where ${\cal U}^* * {\cal S}^* * {{\cal V}^*}^\top$ denotes the t-SVD of $\left({\cal X}^*- \frac{1}{\beta }{{{\cal Z}}^*}\right)$, and $\bar{\cal D}^* = S_p^\mu \left({\bar{\cal S}^*}\right)$. Let $F\left( {{\cal X}} \right) \equiv \frac{1}{2} \left\| {{{{\cal P}}_\Omega }\left( {{{\cal X}} - {{\cal T}}} \right)} \right\|_F^2 + \lambda {\left\| {{\cal X}} \right\|_p}$, the following lemma shows that $F\left({\cal X}\right)$ is always non-increasing ($\beta > 0$) as the iterations proceed.
\\\\
\textbf{Lemma 1.} According to the properties of the $p$-TNN, and when $-\infty < p < 1$, let $\left\{{\cal X}^t\right\}$ be the sequence generated by Algorithm 1. Then, we have the following inequality:
\begin{equation}
F\left( {{{{\cal X}}^{t + 1}}} \right) \le F\left( {{{{\cal X}}^t}} \right) - \frac{\beta }{2}\left\| {{{{\cal X}}^{t + 1}} - {{{\cal X}}^t}} \right\|_F^2.\label{eq29}
\end{equation}
\\
\textbf{Theorem 2.} Let $\left\{{\cal X}^t\right\}$ be the sequence produced by algorithm 1, we say $\left\{{\cal X}^t\right\}$ is a bounded iterative sequence, i.e., $\sum\nolimits_{t = 1}^\infty  {\left\| {{{\cal X}^{t + 1}} - {{\cal X}^t}} \right\|} _F^2 < \infty $.

According to Theorem 2, it can be seen that the proposed algorithm generates a bounded iterative sequence. Moreover, from the above theorem, we must have $\mathop {\lim }\limits_{t \to \infty } \left\| {{{\cal{X}}^{t + 1}} - {{\cal{X}}^t}} \right\|_F^2 = 0$. Hence, the iteration sequence has limit point.

In \eqref{eq19}, we choose $\frac{1}{2}\left\| {{{\cal P}_\Omega }\left( {{\cal X} - {\cal T}} \right)} \right\|_F^2$, which is commonly used in LRTC problem and low-rank matrix completion problem, as the loss function. Particularly, the loss function we choose is $\rho$-Lipschitz smooth, i.e., it satisfies $\left\| {\nabla \ell \left( {{{\cal{X}}_1}} \right) - \nabla \ell \left( {{{\cal{X}}_2}} \right)} \right\|_F^2 \le \rho \left\| {{{\cal{X}}_1} - {{\cal{X}}_2}} \right\|_F^2$. Similar with \cite{b37}, we use ${\left\| {{{\cal X}^{t + 1}} - {{\cal X}^t}} \right\|} _F^2$ to perform the convergence analysis of the proposed algorithm. The convergence of Algorithm 1 is shown in the following theorem, and the proof can be founded in Appendix.
\\\\
\textbf{Theorem 3.} Let $\left\{{\cal X}^t\right\}$ be the sequence produced by Algorithm 1. For the consecutive elements ${\cal X}^t$ and ${\cal X}^{t+1}$, we have
\begin{equation}
\mathop {\min }\limits_{t = 1, \cdots, T} \left\| {{{\cal X}^{t + 1}} - {{\cal X}^t}} \right\|_F^2 \le \frac{2}{{\beta T}}\left( {F\left( {{{{\cal X}}^1}} \right) - \inf F} \right).\label{eq30}
\end{equation}
where ${\rm inf} {\kern 1pt} F$ denotes the minimizer of the objective function $F$. The above theorem shows that Algorithm 1 converge to a critical point at the rate of ${\cal O}\left(1/T\right)$.

\section{Experiments}
In this section, we perform numerical experiments on synthetic and real-world data sets to demonstrate the effectiveness of our proposed algorithm. Furthermore, the experimental results show the superiority of our method. Each experiment is repeated ten times, and the average results are reported. All experiments are implemented in Matlab on Windows 10 with Intel Xeon 2.8GHz CPU and 128GB memory. 
\\
\subsection{Evaluation metrics}

Let ${\cal X},{{\cal{X}}^*} \in {{\mathbb{R}}^{{I_1} \times {I_2} \times {I_3}}}$ denote the underlying tensor and its recovery tensor (i.e., the output of the LRTC methods), respectively. The following metrics are chosen to evaluate the recovery performance of the LRTC algorithms.
\\
1. Relative Square Error, denoted by RSE, is defined as:
\begin{equation}
{\rm{RSE}} = \frac{{{{\left\| {{{\cal X}^*} - {\cal X}} \right\|}_F}}}{{{{\left\| {\cal X} \right\|}_F}}}.\label{eq31}
\end{equation}
\\
2. Peak Signal-to-Noise Ratio, denoted by PSNR, is defined as:
\begin{equation}
{\rm RSNR} = 10{\log _{10}}\frac{{\left\| {\cal X} \right\|_{\max }^2}}{{{{\left\| {{{\cal X}^*} - {\cal X}} \right\|_F^2} \mathord{\left/{\vphantom {{\left\| {{{\cal X}^*} - {\cal X}} \right\|_F^2} {\left( {{I_1}{I_2}{I_3}} \right)}}} \right.
				\kern-\nulldelimiterspace} {\left( {{I_1}{I_2}{I_3}} \right)}}}}.\label{eq32}
\end{equation}
\\
3. Additionally, for the experiments on real-world data sets, another metric, SSIM, is defined as:
\begin{equation}
{\rm SSIM} = \frac{{\left( {2{\mu _{\mathbf X}}{\mu _{{{\mathbf X}^*}}} + {c_1}} \right)\left( {2{\sigma _{{\mathbf X}{{\mathbf X}^*}}} + {c_2}} \right)}}{{\left( {\mu _{\mathbf X}^2 + \mu _{{{\mathbf X}^*}}^2 + {c_1}} \right)\left( {\sigma _{\mathbf X}^2 + \sigma _{{{\mathbf X}^*}}^2 + {c_2}} \right)}}.\label{eq33}
\end{equation}
where ${\mathbf X}$ and ${\mathbf X}^*$ denote the greyscale images for the original image and its recovery image, $c_1$ and $c_2$ are constants, $\mu_{\mathbf X}$ and ${\mu _{{{\mathbf X}^*}}}$ denote the average values, while $\sigma_{\mathbf X}$ and ${\sigma_{{{\mathbf X}^*}}}$ denote the standard deviation of ${\mathbf X}$ and ${\mathbf X}^*$, respectively, and ${{\sigma _{{\mathbf X}{{\mathbf X}^*}}}}$ denote the covariance matrix between ${\mathbf X}$ and ${\mathbf X}^*$.
\\
\subsection{Parameter settings}

The stopping criterion of the proposed algorithm is:
\begin{equation}
t > T,{{{{{\left\| {{{\cal X}^{t + 1}} - {{\cal X}^t}} \right\|}_F}} \mathord{\left/
			{\vphantom {{{{\left\| {{{\cal X}^{t + 1}} - {{\cal X}^t}} \right\|}_F}} {\left\| {{{\cal X}^t}} \right\|}}} \right.
			\kern-\nulldelimiterspace} {\left\| {{{\cal X}^t}} \right\|}}_F} \le tol.\label{eq34}
\end{equation}
where the maximum number of iterations $T$ is fixed at 1000, and the tolerance $tol$ is fixed at $10^{-4}$. Moreover, with regard to the penalty parameter, it is initialized as 0.01, i.e., ${\beta ^0} = 0.01$, the step size is set to 1.1, and ${\beta ^{\max }} = {10^5}$. Additionally, following the settings in \cite{b37}, $\gamma_1 = 0.1$, and $\rho = 2$.
\begin{figure}[h]
	\centering
	\includegraphics[width=0.5\textwidth]{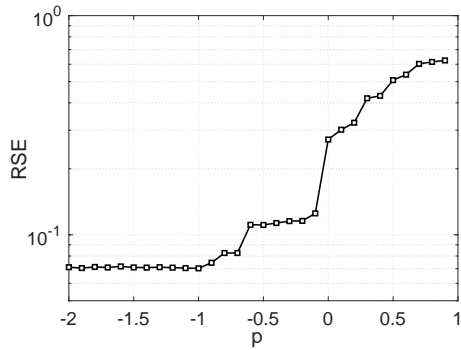}
	\caption{Performance of the proposed method with different selection of $p$-values.}
	\label{Fig.2}
\end{figure}

Furthermore, we investigate the effect of the $p$-value selection on the performance of the proposed method. For a given $100 \times 100 \times 20$ tensor with the tensor tubal rank 5 (which can be generated with the method in synthetic data), we attempt to recover it from $20\%$ sampling observations (i.e., the sampling rate sr = 0.2) to show the performance of our method with different $p$-values. In the experiments, we follow the above parameter settings and vary $p$ in the range $[-2, 0.9]$, and the step size is fixed at 0.1. Fig.\ref{Fig.2} shows the performance of the proposed method with different $p$-values.

As can be seen from Fig.\ref{Fig.2}, the performance tends to be stable when $p \le -1$. Meanwhile, the optimal performance is achieved. Therefore, we fix $p = -1$ in the rest of experiments.
\subsection{Synthetic data}
\textbf{Data sets}: As in \cite{b6,b34}, we assume that the tensor ${\cal X} \in {{\mathbb{R}}^{{I_1} \times {I_2} \times {I_3}}}$ with tubal rank $r$ can be generated by a tensor product ${\cal X} = {\cal P}*{\cal Q}$, where ${\cal P} \in {{\mathbb{R}}^{{I_1} \times {r} \times {I_3}}}$ and ${\cal Q} \in {{\mathbb{R}}^{{r} \times {I_2} \times {I_3}}}$ are tensors with elements obtained independently from the ${\cal N}\left(0,1\right)$ distribution. Three kinds of data sets are generated:

(i) We fix the tubal rank at 5 and consider the tensor ${\cal X} \in {{\mathbb{R}}^{{I_1} \times {I_2} \times {I_3}}}$ with $I_1=I_2=I_3=n$ that $n$ varies in $\left\{50, 100, 150, 200\right\}$. 

(ii) We fix $I=100$ and consider the tensor ${\cal X} \in {{\mathbb{R}}^{{I} \times {I} \times {I_3}}}$ with $I_3$ varies in $\left\{20, 40, 60, 80, 100\right\}$. Moreover, the tensor tubal rank is also fixed at 5. 

(iii) We consider the tensor ${\cal X} \in {{\mathbb{R}}^{{I} \times {I} \times {I_3}}}$ with $I_3$ ($I=100 , I_3= 20$) with different tubal rank $r$ varying in the range $[10, 40]$.
\begin{figure}[h]
	\centering
	\subfigure[]{
		\includegraphics[width=0.45\textwidth]{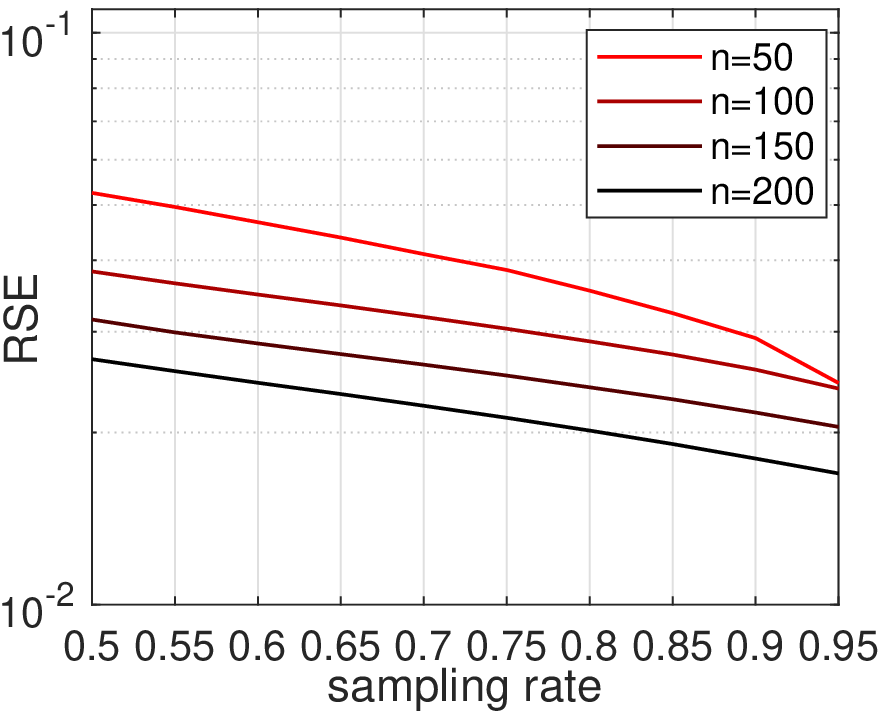}
		\label{subfig1}}
	\subfigure[]{
		\includegraphics[width=0.45\textwidth]{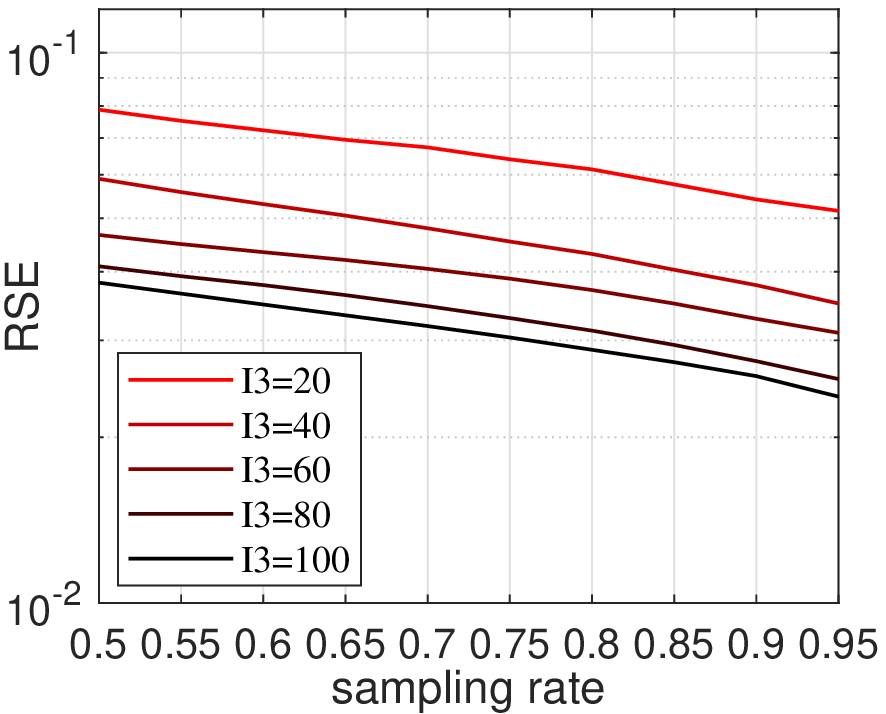}
		\label{subfig2}}
	\caption{Performance of the proposed method on the synthetic data sets. (a) RSEs vs sampling rate on different size tensors. (b) RSEs vs sampling rate on tensors with different $I_3$ }
	\label{Fig.3}
\end{figure}

Considering the first data set, we investigate the effect of different tensor size on the performance of the proposed method. The sampling rate is varied from 0.05 to 0.5, with a step size equaling 0.05. Results are shown in Fig.\ref{Fig.3}(a). Note that there exists a negative correlation between the tensor size and RSE. In particular, the underlying tensor can be recovered accurately with a much small sampling rate, when its size is large enough. In addition, for a fixed-size tensor, the RSE decreases monotonously with an increasing sampling rate. It is reasonable -- the larger the number of observations is, the more information of the underlying tensor is obtained. 

Furthermore, we study the influence of different $I_3$ on the recovery performance using the second data set. The sampling rate is varied from 0.05 to 0.5, with a step size equaling 0.05. Results are shown in Fig.\ref{Fig.3}(b). At the same sampling rate, it can be seen that a larger $I_3$ leads to a smaller RSE. This conclusion is similar to the discussion of small $I_3$ and large $I_3$ in \cite{b38}. Moreover, with an increasing sampling rate, RSEs decreases monotonously. 

\begin{figure}[h]
	\centering
	\includegraphics[width=0.5\textwidth]{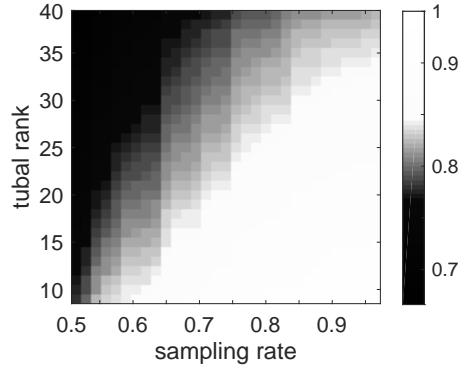}
	\caption{Performance of our method with varying tubal rank and sampling rate. The numbers plotted on the above figure are fraction of successful recoveries within 10 random trials. The white and black areas means ``succeed'' and ``fail'', respectively. Here, the threshold between the two states is set to ${\rm PSNR} = 32dB$. }
	\label{Fig.4}
\end{figure}

Considering the third data set, we investigate the effect of tubal rank on the recovery performance of the proposed method. The sampling rate is varied from 0.05 to 0.5, with a step size equaling 0.015. Furthermore, the PSNRs are normalized. Results are shown in Fig.\ref{Fig.4}, where white means the underlying tensor is recovered successfully while black indicates that the recovery failed. As can be seen, the recovery performance can be guaranteed as long as the tubal rank is relatively low and the sampling rate is relatively large.

\subsection{Color image data}
\textbf{Data sets}: Zhou et al. \cite{b30} point out that most natural images have low tubal rank structure, and following the experiment settings in \cite{b39} and \cite{b40}, we use four color images \footnote {https://github.com/Spring-Liu/testimage.} to demonstrate the performance of the proposed method in this part. The four images (named lena, facade, baboon, and house, respectively) are resized to $256\times256\times3$ tensor. A summary of these four images is reported in Fig.\ref{Fig.5}.

\textbf{Baseline}: We compare the performance of the proposed algorithm with other state-of-the-art methods, including: i) smooth PARAFAC tensor completion (SPC) \cite{b40} which is CP-rank based method, ii) the tensor n-rank based methods, i.e., simultaneous tensor decomposition and completion (STDC) \cite{b25}, iii) the TT rank based methods, i.e., Tmac-TT \cite{b26} and tensor train stochastic gradient descent (TT-SGD) \cite{b34}, and iv) TNN \cite{b33} which is the tubal rank based method.

\begin{table}[]
	\centering
	\caption{Comparison of the recovered results on color image data sets with different sampling rates. The optimal values of PSNR, RSE, and SSIM for each test are highlighted in bold.}
	\setlength{\tabcolsep}{0.1mm}{
	\begin{tabular}{llllllllllllll}
		\hline
		\multicolumn{1}{c}{\multirow{2}{*}{sr}}     & 
		\multicolumn{1}{c}{\multirow{2}{*}{method}} & 
		\multicolumn{3}{c}{lena}                    & 
		\multicolumn{3}{c}{facade}                  & 
		\multicolumn{3}{l}{baboon}                  & 
		\multicolumn{3}{l}{house}                   \\ \cline{3-14} 
		\multicolumn{1}{c}{}                        & 
		\multicolumn{1}{c}{}                        & 
		\multicolumn{1}{c}{PSNR}   & RSE    & SSIM  & PSNR    & RSE    & SSIM   & PSNR     & RSE    
		& SSIM   & PSNR     & RSE  & SSIM   \\ \hline
		\multicolumn{1}{c}{\multirow{6}{*}{0.1}}    
		& \multicolumn{1}{c}{SPC}                     
		& 25.5846 & 0.0948 & 0.7742 
		& \textbf{26.4615} & \textbf{0.0921} & \textbf{0.8189}
		& 21.0733 & 0.1638 & 0.5082 
		& 25.8000 & 0.0842 & 0.7979 \\
		                                                                
		&\multicolumn{1}{c}{STDC}                             
		&19.7222 &0.1862	&0.4401	
		&20.0552 &0.1925	&0.6117	
		&16.9850 &0.2622	&0.3892	
		&19.7105 &0.1697    &0.4099\\
		&\multicolumn{1}{c}{Tmac-TT}                          
		&23.2412 &0.1242	&0.6972	
		&19.7929 &0.1985	&0.6098	
		&18.8118	&0.2096	&0.4046	
		&23.1784	&0.1145	&0.6474\\
		&\multicolumn{1}{c}{TT-SGD}                           
		&15.3518	&0.2282	&0.4245	
		&20.8763	&0.1731	&0.6134	
		&16.0909	&0.2946	&0.3520	
		&16.8428	&0.2448	&0.3781\\
		& \multicolumn{1}{c}{TNN}                
		&24.1932	&0.1080	&0.7105
		&22.1981	&0.1505	&0.6571	 
		&20.6019 &0.1729	&0.4156
		&22.3312 &0.1255    &0.6060    \\ 
		&\multicolumn{1}{c}{proposed}                         
		&\textbf{25.9028}	&\textbf{0.0916}    &\textbf{0.8015}
		&24.3383	&0.1180	&0.7328	
		&\textbf{21.3046}	&\textbf{0.1595}	&\textbf{0.5145}	
		&\textbf{25.9565}	&\textbf{0.0828}	&\textbf{0.8152}\\ \hline
		\multirow{6}{*}{0.2}                       &\multicolumn{1}{c}{SPC}                              
		&27.4107	&0.0768	&0.8474	
		&26.7079	&0.0891	&0.8740	
		&22.2023	&0.1438	&0.6126	
		&27.5046	&0.0692	&0.8480\\
		
		&\multicolumn{1}{c}{STDC}                             
		&25.8928	&0.0915	&0.8066	
		&24.4484	&0.1161	&0.7727	
		&19.6754	&0.1924	&0.4094	
		&28.1505	&0.0642	&0.8673\\
		&\multicolumn{1}{c}{Tmac-TT} 
		&25.8833	&0.0916	&0.8015	
		&22.0045	&0.1538	&0.6506	
		&21.5952	&0.1542	&0.5884	
		&25.9585	&0.0826	&0.8230\\
		& \multicolumn{1}{c}{TT-SGD} 
		&20.8364	&0.1677	&0.6224	
		&25.2864	&0.1067	&0.7953	
		&18.7745	&0.2125	&0.3959	
		&21.4953	&0.1363	&0.4762\\
		&\multicolumn{1}{c}{TNN}
		&26.0373	&0.0900	&0.8151	
		&26.4776	&0.0919	&0.8213	
		&21.9099	&0.1487	&0.6032	
		&25.5434	&0.0867	&0.7918\\
		&\multicolumn{1}{c}{proposed}                         
		&\textbf{28.4658}	&\textbf{0.0681}	&\textbf{0.8820}
		&\textbf{27.9356}	&\textbf{0.0783}	&\textbf{0.8790}	
		&\textbf{22.7415}	&\textbf{0.1352}	&\textbf{0.6565}	
		&\textbf{29.3077}	&\textbf{0.0566}	&\textbf{0.8894}\\ \hline
		\multirow{6}{*}{0.3}                     
		&\multicolumn{1}{c}{SPC}
		&28.7736	&0.0657	&0.8871	
		&28.6894	&0.0713	&0.8927	
		&23.2435	&0.1276	&0.7150	
		&29.0217	&0.0581	&0.8828\\
		
		&\multicolumn{1}{c}{STDC}                             
		&27.9464	&0.0722	&0.8553	
		&25.6342	&0.1013	&0.8092	
		&21.7230	&0.1520	&0.5933	
		&30.4959	&0.0490	&0.9129\\
		&\multicolumn{1}{c}{Tmac-TT}                          
		&26.8824	&0.0816	&0.8341	
		&23.6913	&0.1267	&0.7285	
		&22.8057	&0.1342	&0.6657	
		&26.9618	&0.0736	&0.8375\\
		&\multicolumn{1}{c}{TT-SGD}                           
		&24.0597	&0.1113	&0.7066	
		&28.1859	&0.0750	&0.8900	
		&20.7091	&0.1709	&0.4982	
		&24.1978	&0.1010	&0.6512\\
		&\multicolumn{1}{c}{TNN}                             
		&28.0295	&0.0715	&0.8649	
		&28.8579	&0.0699	&0.9100	
		&22.9498	&0.1320	&0.6867	
		&27.8168	&0.0667	&0.8652\\
		&\multicolumn{1}{c}{proposed}                         
		&\textbf{30.4150}	&\textbf{0.0545}	&\textbf{0.9324}
		&\textbf{30.1530}	&\textbf{0.0602}	&\textbf{0.9270}	
		&\textbf{24.0340}	&\textbf{0.1165}	&\textbf{0.7626}	
		&\textbf{32.1177}	&\textbf{0.0407}	&\textbf{0.9225}\\ \hline
		\multirow{6}{*}{0.4}                     	&\multicolumn{1}{c}{SPC}                              
		&30.0387	&0.0568	&0.9147	
		&29.0100	&0.0687	&0.9142	
		&24.2049	&0.1142	&0.7897	
		&29.8044	&0.0531	&0.9070\\
		
		&\multicolumn{1}{c}{STDC}                             
		&29.1445	&0.0629	&0.8960	
		&26.6845	&0.0898	&0.8721
		&23.4226	&0.1250	&0.7475	
		&31.5988	&0.0432	&0.9180\\
		&\multicolumn{1}{c}{Tmac-TT}  
		&27.8189	&0.0733	&0.8511	
		&25.5229	&0.1026	&0.8081
		&23.7994	&0.1197	&0.7497	
		&27.9107	&0.0660	&0.8663\\
		&\multicolumn{1}{c}{TT-SGD}                           
		&26.6475	&0.0856	&0.8272	
		&29.6810	&0.0619	&0.9248	
		&22.3022	&0.1412	&0.6204	
		&25.8701	&0.0831	&0.8015\\
		&\multicolumn{1}{c}{TNN}                             
		&29.8159	&0.0582	&0.9005	
		&30.4659	&0.0581	&0.9337		
		&23.9500	&0.1176	&0.7580	
		&29.4037	&0.0556	&0.8934\\
		&\multicolumn{1}{c}{proposed}                         
		&\textbf{32.5232}	&\textbf{0.0427}	&\textbf{0.9582}
		&\textbf{31.3102}	&\textbf{0.0527}	&\textbf{0.9466}	
		&\textbf{25.3032}	&\textbf{0.1006}	&\textbf{0.8308}	
		&\textbf{33.6028}	&\textbf{0.0343}	&\textbf{0.9417}\\ \hline
	\end{tabular}}
\end{table}
For each color image, we test the above mentioned LRTC methods with sampling rate equaling 0.1, 0.2, 0.3 and 0.4. Results are shown in Table 1. As can be seen from Table 1, the performance improvement is universal for an increasing sampling rate. In particular, our proposed method is generally superior to other state-of-the-arts. More precisely, in addition to the results of facade at sampling rate 0.1, our method achieves highest PSNR, SSIM and smallest RSE. For further visually compare the recovery performance of all methods, we show the recovered results for each color image in Fig.5 (the sampling rate is fixed at 0.2). Note that the recovered color images of our method are closer to the original images. The reason of our method can obtain better details of images than TNN is that the proposed $p$-TNN is a much tighter approximation of the tubal rank than tensor nuclear norm.

\subsection{Hyperspectral image data}
\textbf{Data sets:} In this part, we use two public hyperspectral image data sets named Washington DC Mall (WDC Mall) and Pavia University (paviaU)\footnote {http://www.ehu.eus/ccwintco/index.php?title=Hyperspectral\_Remote\_Sensing\_Scenes.}. The whole WDC Mall data set contains $1208 \times 307$ pixels and 191 spectral bands, which is from the hyperspectral digital imagery collection experiment. In the following experiments, it is resized to a $200 \times 200 \times 191$ tensor. The whole paviaU data set is a $610 \times 610$ pixels and 103 spectral bands image, which is from Pavia University. In the rest experiments, it is resized to a $200 \times 200 \times 103$ tensor. A visually summary is shown in Fig. 6.
\\
\textbf{Baseline:} We compare the performance of the proposed algorithm with the same methods mentioned in color image experiments, i.e., SPC, STDC, Tmac-TT, TT-SGD, and TNN.

\begin{table}[]
	\centering
	\caption{Comparison of the recovered results on Hyperspectral image data sets with different sampling rates. The optimal values of PSNR and RSE for each test are highlighted in bold.}
	\setlength{\tabcolsep}{1mm}{
	\begin{tabular}{llllll}
		\hline
		\multicolumn{1}{c}{\multirow{2}{*}{sr}}     & 
		\multicolumn{1}{c}{\multirow{2}{*}{method}} & 
		\multicolumn{2}{c}{WDC Mall}                & 
		\multicolumn{2}{c}{PaviaU}                  \\ \cline{3-6} 
		\multicolumn{1}{c}{}                        & 
		\multicolumn{1}{c}{}                        & 
		\multicolumn{1}{c}{PSNR}   & RSE  & PSNR    & RSE    \\ \hline
		\multicolumn{1}{c}{\multirow{6}{*}{0.1}}    
		& \multicolumn{1}{c}{SPC}
		&22.9355	&0.1926
		&26.1087	&0.2270
		\\
		                                                    
		&\multicolumn{1}{c}{STDC}                             
		&16.1068	&0.4221	
		&21.4573	&0.3877
		\\
		&\multicolumn{1}{c}{Tmac-TT}                          
		&16.5548	&0.4009	
		&21.6522	&0.3791	
		\\
		&\multicolumn{1}{c}{TT-SGD}                           
		&17.4878	&0.3606	
		&21.0885	&0.4045	
		\\
		& \multicolumn{1}{c}{TNN}                
		
		&19.6377	&0.2813	
		&23.2825	&0.3142	 
		\\
		&\multicolumn{1}{c}{proposed}                         
		&\textbf{22.9442}	&\textbf{0.1924}
		&\textbf{27.3111}	&\textbf{0.1976}
		\\ \hline
		\multirow{6}{*}{0.2}                       &\multicolumn{1}{c}{SPC}                              
		&26.8823	&0.1222	
		&31.2002	&0.1263
		\\
		
		&\multicolumn{1}{c}{STDC}                             
		&20.8547	&0.2447	
		&24.2979	&0.2796	
		\\
		&\multicolumn{1}{c}{Tmac-TT} 
		&18.7733	&0.3105
		&23.4255	&0.3091	
		\\
		& \multicolumn{1}{c}{TT-SGD} 
		&20.0943	&0.2668	
		&24.7403	&0.2657	
		\\
		&\multicolumn{1}{c}{TNN}
			
		&21.1869	&0.2353	
		&24.9034	&0.2608
		\\
		&\multicolumn{1}{c}{proposed}                         
		&\textbf{27.6564}	&\textbf{0.1118}	
		&\textbf{31.4136}	&\textbf{0.1234}	
		\\ \hline
		\multirow{6}{*}{0.3}                     
		&\multicolumn{1}{c}{SPC}
		&29.8279	&0.0871	
		&34.2700	&0.0887	
		\\
		
		&\multicolumn{1}{c}{STDC}                             
		&22.8627	&0.1941	
		&26.4978	&0.2171	
		\\
		&\multicolumn{1}{c}{Tmac-TT}                          
		&20.5384	&0.2534	
		&25.0036	&0.2577	
		\\
		&\multicolumn{1}{c}{TT-SGD}                           
		&21.7456	&0.2206	
		&26.2153	&0.2242	
		\\
		&\multicolumn{1}{c}{TNN}                             
		&24.5025	&0.1608	
		&27.7402	&0.1881
		\\
		&\multicolumn{1}{c}{proposed}                         
		&\textbf{30.0019}	&\textbf{0.0854}	
		&\textbf{34.4901}	&\textbf{0.0865}	
		\\ \hline
		\multirow{6}{*}{0.4}                     	&\multicolumn{1}{c}{SPC}                              
		&31.1864	&0.0745	
		&35.6849	&0.0754
		\\
		
		&\multicolumn{1}{c}{STDC}                             
		&24.2289	&0.1658	
		&28.0926	&0.1807
		\\
		&\multicolumn{1}{c}{Tmac-TT}  
		&22.0513	&0.2129
		&26.6126	&0.2142
		\\
		&\multicolumn{1}{c}{TT-SGD}                           
		&23.2092	&0.1863	
		&27.5421	&0.1924
		\\
		&\multicolumn{1}{c}{TNN}                             
		&26.6422	&0.1257	
		&29.6721	&0.1507
		\\
		&\multicolumn{1}{c}{proposed}                         
		&\textbf{32.1133}	&\textbf{0.0669}	
		&\textbf{36.9243}	&\textbf{0.0653}	
		\\ \hline
	\end{tabular}}
\end{table}

For the two hyperspectral images, we test the above mentioned LRTC methods with sampling rate equaling 0.1, 0.2, 0.3 and 0.4. Results are shown in Table 2. We use only PSNR and RSE as the evaluation metrics here. Similarly, we can see that our method almost outperforms others. The improvement of average recovery performance is statistically significant. To further prove the superiority of our proposed method for hyperspectral image inpainting, we show the recovered results of all methods from 0.2 sampling observed images in Fig.6, where only three slices of two data sets (the first three slices of WDC Mall and slices [60,61,62] of paviaU) are shown exemplarily. As can be seen from Fig.6, compared with other methods, our proposed algorithm visually performs better recovery results.

\subsection{Conclusion}
In this paper, we address the problem of low-rank tensor completion. We propose a new definition of tensor $p$-shrinkage nuclear norm ($p$-TNN). The tightness property of $p$-TNN demonstrates that the proposed $p$-TNN is a tighter surrogate of tensor average rank than tensor nuclear norm (TNN). Therefore, we propose a novel LRTC model by employing our proposed $p$-TNN. In particular, the upper bound of recovery error for our LRTC model is further provided to show the underlying tensor can be recovered accurately. Accordingly, we develop an efficient algorithm by incorporating the adaptive momentum scheme. Subsequently, some theoretical analysis are provided from the aspects of complexity and convergence, respectively. The experimental results validate the superiority of our method over the state-of-the-arts.

However, there still several directions in future work. First, the experimental results show that further improvements are needed for the performance of our method at low sampling rate. Moreover, we will focus on the distributed version to apply it to massive data sets.

\end{document}